# MOMENTSNET: A SIMPLE LEARNING-FREE METHOD FOR BINARY IMAGE RECOGNITION

*Jiasong Wu*[1, 2, 3, 4*], *Shijie Qiu*[1, 4], *Youyong Kong*[1, 4], *Yang Chen*[1, 4], *Lotfi Senhadji*[2, 3, 4], *Huazhong Shu*[1, 4]

[1]LIST, the Key Laboratory of Computer Network and Information Integration (Southeast University), Ministry of Education, 210096 Nanjing, China
[2]INSERM, U1099, Rennes, F-35000, France
[3]Université de Rennes 1, LTSI, Rennes, F-35042, France
[4]Centre de Recherche en Information Médicale Sino-français (CRIBs)
Email: jswu@seu.edu.cn, 1063392760@qq.com, kongyouyong@seu.edu.cn, chenyang.list@seu.edu.cn, lotfi.senhadji@univ-rennes1.fr, shu.list@seu.edu.cn

## ABSTRACT

In this paper, we propose a new simple and learning-free deep learning network named MomentsNet, whose convolution layer, nonlinear processing layer and pooling layer are constructed by Moments kernels, binary hashing and block-wise histogram, respectively. Twelve typical moments (including geometrical moment, Zernike moment, Tchebichef moment, etc.) are used to construct the MomentsNet whose recognition performance for binary image is studied. The results reveal that MomentsNet has better recognition performance than its corresponding moments in almost all cases and ZernikeNet achieves the best recognition performance among MomentsNet constructed by twelve moments. ZernikeNet also shows better recognition performance on binary image database than that of PCANet, which is a learning-based deep learning network.

## 1. INTRODUCTION

Convolutional neural networks (CNNs) with backpropagation process have shown its great success in various image classification tasks [1-6]. However, backpropagation process has very high computational complexity. To address this issue, convolutional networks without backpropagation have been proposed by many researchers. Chan *et al*. [7] proposed a lightweight and learning-based convolutional network named principle component analysis network (PCANet), who extracts the feature of the input images by using cascaded PCA filter, binary hashing, and block-wise histogram successively. PCANet does not use the backpropagation process but works unexpectedly well in most image classification tasks. Due to the success of PCANet, many researchers try to modify and extend this network in recent years [8-12]. Ng and Teoh [8] proposed a discrete cosine transform network (DCTNet) for face recognition. Feng et al. [9] proposed a discriminative locality alignment network (DLANet) for scene classification. Yang et al. [10] proposed a canonical correlation analysis network (CCANet) for two-view image recognition.

On the other hand, image moments [13-23] are widely used in binary image recognition due to their invariant representations of images. Geometric moments are the simplest moments but they are not orthogonal. Teague [13] introduced the continuous orthogonal moments (COTs) for which Legendre moments and Zernike moments are two typical representatives. To overcome the discretization error problem of the COTs, researchers proposed various discrete orthogonal moments, including Tchebichef moments [14], Krawtchouk moments [15] and dual Hahn moments [16], etc. In order to get a more invariant image representation, Yap et al. [17] produced a set of polar harmonic transforms (PHTs), including polar complex exponential transform (PCET), polar cosine transform (PCT), and polar sine transform (PST). Hoang and Tabbone [18] then presented a set of generic PHTs (GPHTs), including generic PCET (GPCET), generic PCT (GPCT), and generic PST (GPST). Note that many fast algorithms [for example, 19-21] are proposed for moments due to their learning-free properties, while deriving the fast algorithms for learning-based PCA is very difficult. For more references on moments, we refer to [24-26].

In this paper, we propose a novel simple and learning-free deep learning network named MomentsNet for extracting the features of the input images. Then, the recognition performance of MomentsNet is analyzed with respect to various parameters, including thresholding in binarization process and also compared to PCANet.

The rest of the paper is organized as follows. In Section 2, various moments are briefly introduced. The architecture of MomentsNet is described in Section 3. The recognition performance of MomentsNet, moments, and PCANet are compared in Section 4. Section 5 concludes the paper.

## 2. PRELIMINARIES

The general $(n+m)$th order two-dimensional (2D) moment definition is given by

$$\Psi_{nm} = \iint_{R^2} \varphi_{nm}(x,y) f(x,y) dx dy$$
$$= \int_0^{2\pi} \int_0^1 \varphi_{nm}(r,\theta) f(r,\theta) r dr d\theta, \quad n,m = 0,1,2,... \quad (1)$$

, where $\varphi_{nm}(x,y)$ and $\varphi_{nm}(r,\theta)$ are moments kernels in Cartesian coordinates and in polar coordinates, respectively. $f(x,y)$ and $f(r,\theta)$ are image values in Cartesian coordinates and in polar coordinates, respectively. A summary of the moments used in this paper is given in Table 1.

## 3. THE ARCHITECTURE OF MOMENTSNET

The architecture of the proposed MomentsNet is summarized in Fig. 1. We then take the MomentsNet-2, in which the first two convolutional layers are constructed by moments, as an example to describe in details.

Table 1. The definitions of various moments

| Moments | Definition | Moments | Definition |
|---|---|---|---|
| Geometric | $G_{nm} = \iint_{\mathbb{R}^2} x^n y^m f(x,y) dx dy$ | Legendre [13] | $L_{nm} = \frac{(2n+1)(2m+1)}{4} \int_{-1}^{1} \int_{-1}^{1} P_n(x) P_m(y) f(x,y) dx dy$, $P_n(x) = \frac{1}{2^n} \sum_{k=0}^{n/2} (-1)^k \frac{(2n-2k)!}{k!(n-k)!(n-2k)!} x^{n-2k}$. |
| Zernike [13] | $Z_{nm} = \frac{n+1}{\pi} \int_0^1 \int_0^{2\pi} E_{nm}(r) e^{-jm\theta} f(r,\theta) r dr d\theta$, $0 \le \|m\| \le n, n-\|m\|$ is even, $E_{nm}(r) = \sum_{k=0}^{(n-\|m\|)/2} (-1)^k \frac{(n-k)!}{k![(n-2k+\|m\|)/2]![(n-2k-\|m\|)/2]!} r^{n-2k}$ | Tchebichef [14] | $D_{nm} = [\rho(n,N)\rho(m,M)]^{-1} \sum_{x=0}^{N-1} \sum_{y=0}^{M-1} t_n(x) t_m(y) f(x,y)$, $t_n(x) = n! \sum_{k=0}^{n} (-1)^{n-k} \binom{N-1-k}{n-k}\binom{n+k}{n}\binom{x}{k}$, $\rho(n,N) = [N(N^2-1)(N^2-2^2) \cdots (N^2-n^2)]/(2n+1)$. |
| Krawtchouk [15] | $Q_{nm} = \sum_{x=0}^{N-1} \sum_{y=0}^{M-1} \bar{K}_n(x;p_1,N-1) \bar{K}_m(y;p_2,M-1) f(x,y)$, $\bar{K}_n(x;p,N) = K_n(x;p,N)\sqrt{w(x;p,N)/\rho(n;p,N)}$, $w(x;p,N) = \binom{N}{x} p^x (1-p)^{N-x}$, $\rho(n;p,N) = \left(\frac{p-1}{p}\right)^n \frac{n!}{(-N)_n}$, $K_n(x;p,N) = {}_2F_1(-n,-x;-N;1/p)$, ${}_2F_1(a,b;c;z) = \sum_{k=0}^{\infty} \frac{(a)_k (b)_k}{(c)_k} \frac{z^k}{k!}$, $(a)_k = a(a+1)(a+2)\ldots(a+k-1) = \Gamma(a+k)/\Gamma(a)$. | Dual Hahn [16] | $W_{nm} = \sum_{x=a}^{b-1} \sum_{y=a}^{b-1} \hat{w}_n^{(c)}(x,a,b) \hat{w}_m^{(c)}(y,a,b) f(x,y), m,n=0,1,2,\ldots,N-1$, $\hat{w}_n^{(c)}(x,a,b) = w_n^{(c)}(x,a,b)\sqrt{\frac{\rho(x)}{d_n^2}[\Delta x(x-\frac{1}{2})]}$, $w_n^{(c)}(x,a,b) = \frac{(a-b+1)_n(a+c+1)_n}{n!}$ $\times {}_3F_2(-n,a-x,a+x+1;a-b+1,a+c+1;1)$, $\rho(x) = \frac{\Gamma(a+x+1)\Gamma(c+x+1)}{\Gamma(x-a+1)\Gamma(b-x)\Gamma(b+x+1)\Gamma(x-c+1)}$, $d_n^2 = \frac{\Gamma(a+c+n+1)}{n!(b-a-n-1)!\Gamma(b-c-n)}$, ${}_3F_2(a_1,a_2,a_3;b_1,b_2;z) = \sum_{k=0}^{\infty} \frac{(a_1)_k (a_2)_k (a_3)_k}{(b_1)_k (b_2)_k} \frac{z^k}{k!}$, $-1/2 < a < b, \|c\| < 1+a, b = a+N$. |
| PCET [17] | $C_{nm} = \int_0^{2\pi} \int_0^1 [H_{nl}(r,\theta)]^* f(r,\theta) r dr d\theta$, $\|n\|=\|m\|=0,1,\cdots,\infty$, $H_{nm}(r,\theta) = R_n(r) e^{im\theta}$, $R_n(r) = e^{i 2\pi n r^2}$, where $[.]^*$ denotes the complex conjugate. | GPCET [18] | $O_{nms} = \int_0^{2\pi} \int_0^1 [V_{nm}(r,\theta)]^* f(r,\theta) r dr d\theta$, $V_{nm}(r,\theta) = R_{ns}(r) e^{im\theta}$, $R_{ns}(r) = \sqrt{\frac{sr^{s-2}}{2\pi}} e^{i 2\pi n r^s}$ |
| PCT [17] | $C_{nm}^C = \int_0^{2\pi} \int_0^1 [H_{nm}^C(r,\theta)]^* f(r,\theta) r dr d\theta$, $n,\|m\|=0,1,\cdots,\infty$, $H_{nm}^C(r,\theta) = R_n^C(r) e^{im\theta}$, $R_n^C(r) = \begin{cases} 1, n=0 \\ \sqrt{2}\cos(\pi n r^2), n>0 \end{cases}$ | GPCT [18] | $O_{nms}^C = \int_0^{2\pi} \int_0^1 [V_{nm}^C(r,\theta)]^* f(r,\theta) r dr d\theta$, $V_{nm}^C(r,\theta) = R_{ns}^C(r) e^{im\theta}$, $R_{ns}^C(r) = \sqrt{\frac{sr^{s-2}}{2\pi}} \begin{cases} 1, n=0 \\ \sqrt{2}\cos(\pi n r^s), n>0 \end{cases}$ |
| PST [17] | $C_{nm}^S = \int_0^{2\pi} \int_0^1 [H_{nm}^S(r,\theta)]^* f(r,\theta) r dr d\theta, n=1,\cdots,\infty, \|m\|=0,1,\cdots,\infty$, $H_{nm}^S(r,\theta) = R_n^S(r) e^{im\theta} = \sin(\pi n r^2) e^{im\theta}$, | GPST [18] | $O_{nms}^S = \int_0^{2\pi} \int_0^1 [V_{nm}^S(r,\theta)]^* f(r,\theta) r dr d\theta$, $V_{nm}^S(r,\theta) = R_{ns}^S(r) e^{im\theta}$, $R_{ns}^S(r) = \sqrt{\frac{sr^{s-2}}{2\pi}} \sqrt{2} \sin(\pi n r^s), n>0$ |

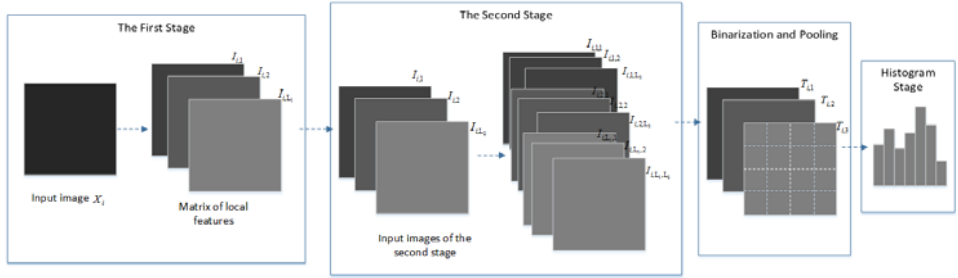

**Fig. 1.** The Architecture of MomentsNet-2

### 3.1. The first stage of MomentsNet

Suppose that we have a set of $N_1$ image samples $X_i \in \mathbb{R}^{M \times N}$, $i=1,2,\ldots,N_1$, for training. We slide these images by using a patch, whose size is $k_1 \times k_2$ and $1 < k_1 < M$, $1 < k_2 < N$. We collect all the overlapping patches around each pixel of each image and then center each patch by subtracting its mean and obtain $I_i = \begin{bmatrix} \bar{x}_{i,1} & \bar{x}_{i,2} & \cdots & \bar{x}_{i,mn} \end{bmatrix} \in \mathbb{R}^{k_1 k_2 \times MN}$ for the *ith* image sample. Then we reshape all the vector $\bar{x}_{i,j} \in \mathbb{R}^{k_1 k_2}$, $j=1, 2, \ldots, MN$, in $I_i$ to the matrix $U_{i,j} \in \mathbb{R}^{k_1 \times k_2}$.

By applying Eq. (1) to $U_{i,j}$, we can get $n_k$ moments coefficients for each patch, where $n_k$ denotes the total number of moments values of orders 0, 1, ..., k. Then combining moments coefficients of each patch in the *ith* image sample to form a matrix $A_i \in \mathbb{R}^{n_k \times MN}$. Let $L_l$ be the desired number of filters in the *lth* layer, we get the first $L_1$ rows of matrix $A_i$ and reshape all the $L_1$ rows to matrices $I_{i,j} \in \mathbb{R}^{M \times N}$, $j=1, 2, \ldots, L_1$. Therefore, {$X_i$, $i=1,2,\ldots,N_1$} is transformed into $N_1 L_1$ moments feature maps {$I_{i,j}$, $i=1,2,\ldots,N_1; j=1,2,\ldots,L_1$} according to the number of filters $L_1$.

It turns out that each element in {$I_{i,j}$, $j=1,2,\ldots,L_1$} can also be used as an input pattern. By repeating the above process, high-level features can be derived.

### 3.2. The second stage of MomentsNet

Assume that we have $L_2$ filters in the second stage. By repeating the same process as the first stage, for each $I_{i,j} \in \mathbb{R}^{M \times N}$, we can get $L_2$ moments feature maps { $I_{i,j,h} \in \mathbb{R}^{M \times N}$, $h=1,2,\ldots,L_2$}. Therefore, {$I_{i,j}$, $i=1,2,\ldots,N_1; j=1,2,\ldots,L_1$} is transformed into $N_1 L_1 L_2$ moments feature maps {$I_{i,j,h}$, $i=1,2,\ldots,N_1; j=1,2,\ldots,L_1$; $h=1,2,\ldots,L_2$} after the second stage of MomentsNet.

### 3.3. The output stage of MomentsNet

The moments feature maps extracted from the second stage are binarized, weighted, and summed to reduce the redundancy features.

Firstly, each element in {$I_{i,j,h}$, $i=1,2,\ldots,N_1; j=1,2,\ldots,L_1$; $h=1,2,\ldots,L_2$} is binarized by using a modified Heaviside function defined as $H(x) = \begin{cases} 0, & x < t \\ 1, & x \geq t \end{cases}$, where $t$ is a threshold. Note that in this step, MoementsNet is different from PCANet which uses the original Heaviside function, that is, $t=0$. Why we use the modified Heaviside function? Because the values of some moments (for example, Zernike moments) are always larger than zero which makes $t=0$ is unsuitable. After this step, we get the binarized features {$J_{i,j,h} \in \mathbb{R}^{M \times N}$, $i=1,2,\ldots,N_1; j=1,2,\ldots,L_1; h=1,2,\ldots,L_2$}.

Since different moment features can capture different variations of the original images, the binarized features should be weighted to form new single moments features as: $T_{i,j} = \sum_{k=1}^{L_2} 2^{k-1} J_{i,j,k}$, $i=1,2,\ldots,N_1; j=1,2,\ldots,L_1$. The pixel values of $T_{i,j}$ are integers in the range of [0, $2^{L_2}-1$].

Next, we use a block of size $h_1 \times h_2$ to slide each of the $L_1$ images $T_{i,j}$, $j=1,2,\ldots,L_1$, with overlap ratio $R$. Then, the features are divided into $B$ boxes and we compute the histogram of the decimal values for each box and denote it as $hist(B)_d$, $d=1,2,\ldots,B$. After this pooling process, we concatenate all the histograms of $B$ boxes into one vector and obtain

$$f_i = [Bhist(T_{i,1}), Bhist(T_{i,2}), \ldots, Bhist(T_{i,L_1})]^T \in \mathbb{R}^{(2^{L_2})L_1 B}$$

The feature vector is then sent to a SVM classifier [27] to get the final recognition results.

## 4. EXPERIMENTAL RESULTS

### 4.1. Experimental database

The database used in the following experiments is a set of binary images of 9 classes, including bird, camel, children, elephant, fork, hammer, key, ray and turtle. Each class contains 144 images and all the images are rotated from 0° to 330° with an increment of 30°. Some image samples are shown in Fig. 2. All the images are then scaled to 32×32 to reduce the computational complexity in the experiments.

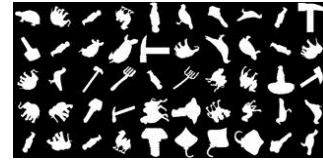

**Fig. 2**. Some Samples of Binary Image Database

We randomly pick up half of the images in each class as the training set and the others left as the testing set.

### 4.2. The impact of various parameters

In this subsection, we focus on the selection of various parameters. We take ZernikeNet as an example to show the impact of the parameters in the following.

*The impact of threshold t:*

From our experiments, the threshold $t$ in modified Heaviside function has a major effect on the final recognition result. To discuss the effect of threshold $t$, we firstly set the number of filters, patch size, block size and overlapping ratio to 9, 11×11, 8×8, and 0.5, respectively. The experiments are carried out on the one-layer Zernike Moments networks (ZernikeNet-1) for simplicity. Fig. 3(a) shows the recognition rates vary with the change of threshold $t$. We can see that there is a dramatic change in the recognition rate as the threshold $t$ changes and the recognition rate is relatively good when $0.1 \leq t \leq 0.2$. A question is raised: *Is there any other parameters that help to choose the threshold value t*? Since $t$ decides the proportion of the number of 1 and 0 in the image after the binarization, we thus additionally record the percentages of the number 1. Fig. 3(b) is the recognition result or the percentage of 1 after binarization vary with the threshold changes from 0.1 to 0.2 with step 0.01. From the figure, we can find that when the percentage of 1 is in the range of [0.4, 0.5], the thresholds correspond to good recognition results. This is in fact a good guidance for the choice of threshold value $t$ not only for ZernikeNet-1 but also for ZernikeNet-2. For ZernikeNet-1, we set the threshold $t$ to 0.1.

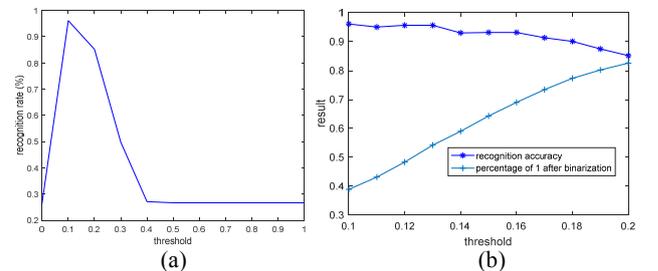

Fig. 3 The impact of threshold $t$. (a) the recognition rate versus the threshold $t$. (b) the "result" versus the threshold $t$. Note that the "result" denotes the recognition rate for the curve above and the percentage of 1 after binarization for the curve below.

*The impact of the number of filters $L_l$, patch size $k_1 \times k_2$, block size $h_1 \times h_2$, and overlapping ratio R:*

Fig. 4 shows the recognition results as the parameters change. It is apparent that the recognition rate tends to be better with the increase of the number of filters $L_1$, but the performance does not improve greatly when $L_1 > 13$. The recognition rate tends to first rise and then fall with the increase of the patch size $k_1 \times k_2$ and the block size $h_1 \times h_2$. ZenikeNet-1 can get the best performance when the patch size and the block size are 11×11 and 4×4, respectively. Comparing with other parameters, the overlapping ratio R has little impact on the recognition rate. We simply set $R=0.5$. ZenikeNet-1 achieves the recognition accuracy of 96.76% with the aforementioned parameter setting.

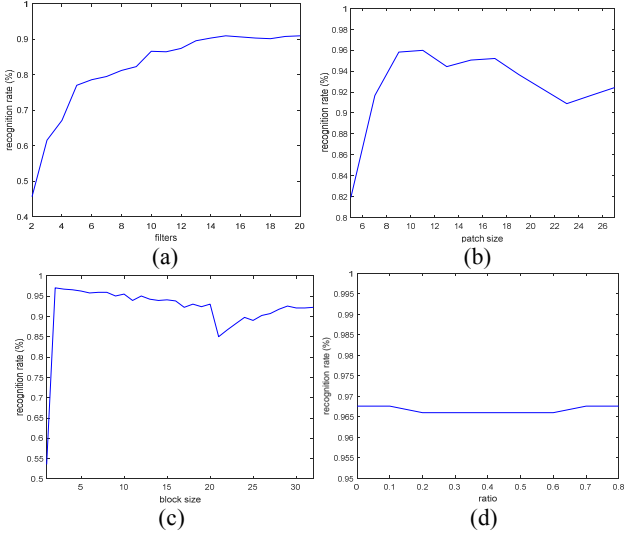

Fig. 4 The recognition rates versus various other parameters. (a) the recognition rate versus the number of filters; (b) the recognition rate versus the patch size; (c) the recognition rate versus the block size; (d) the recognition rate versus the overlap ratio.

### 4.3. Experimental results

We compare the performance of the proposed MomentsNet with the moments and also the PCANet [7]. The best recognition rates and their corresponding parameter settings are shown in Table 2. Note that the accuracy of moments is obtained by extracting the features by using various moments and then feed the features into the SVM classifier [27].

From Table 2, we can see that MomentsNet has better recognition performance than its corresponding moments in almost all cases and ZernikeNet performs the best and GeometryNet performs the worst in terms of the recognition performance among MomentsNet constructed by twelve moments. Most learning-free MomentsNet shows better recognition performance than that of PCANet [7], which is a learning-based deep learning network. We also find that the recognition performance of some MomentsNet-1 outperforms its corresponding MomentsNet-2, for example, TchebichefNet. We think the reason is that the descriptive powers of these moments tend to its upper limit when they are used in the construction of one-stage network, therefore, adding more stage is not helpful for improving the recognition performance.

Table 2. The best recognition rates of different methods. Quintets ($L_1$, $k_1$, $h_1$, $R$, $t$) whose elements denote the number of filters, patch size, block size, overlap ratio, threshold, respectively. Note that we always set $L_2=L_1$, $k_2=k_1$, and $h_2=h_1$ in this paper.

| Moments | The accuracy of Moments (%) | The accuracy of MomentsNet-1 (%) | The accuracy of MomentsNet-2 (%) |
|---|---|---|---|
| Geometry | 13.12 ($n+m$=4) | 73.61 (9, 13, 3, 0.5, 0) | 73.15 (9, 13, 3, 0.5, 0) |
| Legendre | 34.10 ($n+m$=16) | 86.11(11,11,3,0.6, -0.02) | 92.13 (11, 11, 3, 0.5, 0) |
| Zernike | 81.79 ($n+m$=16) | **96.76**(10, 11, 2, 0.5, 0) | **97.69**(10, 11, 2, 0.5, 0) |
| Tchebichef | 62.35 ($n+m$=17) | 93.52(12, 13, 3, 0.4, 0.1) | 89.81(12, 13, 3, 0.5, 0) |
| Krawtchouk | 58.18 ($n+m$=15) | 90.74(8, 13, 4, 0.5, 0) | 93.98(8, 13, 4, 0.5, 0.1) |
| Dual Hahn | 58.02 ($n+m$=15) | 94.75(12, 13, 3, 0.5, 0) | 95.99(12, 15, 4, 0.5, 0) |
| PCT | 93.06 ($n+m$=15) | 94.29(13,15,4, 0.7, 0.01) | 93.36(13,15,4,0.5, 0.006) |
| PCET | 86.27 ($n+m$=16) | 96.45(8, 13, 5, 0.7, 0.04) | 96.91(8, 13, 5, 0.5, 0.002) |
| PST | 92.44 ($n+m$=14) | 91.67(8, 13, 3, 0.6, 0.02) | 88.12(8, 13, 3, 0.5, 0.0003) |
| GPCT | 95.68 ($n+m$=17) | 95.22(8, 11, 6, 0.6, 0.1) | 93.98(8, 11, 6, 0.5, 0.01) |
| GPCET | 92.90 ($n+m$=15) | 95.52(8, 15, 6, 0.8, 0.07) | 96.45(8, 15, 6, 0.5, 0.01) |
| GPST | 95.99 ($n+m$=15) | 96.30(12, 15, 3, 0.6, 0.1) | 96.30(12,15,3, 0.5, 0.012) |
| PCANet [7] | 61.88(325 principal components) | 91.98(8, 7, 7, 0.5, 0) | 94.60(8, 9, 8, 0.4, 0) |

### 5. CONCLUSION

In this paper, we propose a simple and learning-free deep learning network named MomentsNet in which 12 different moments are explored in detail. The impact of various parameters on the recognition performance is analyzed in MomentsNet. MomentsNet has better recognition performance than its corresponding moments in almost all cases and also shows better recognition performance than that of PCANet in binary image recognition.

### ACKNOWLEDGEMENT

This work was supported by the National Natural Science Foundation of China (No. 61201344, 61271312, 61401085, 31640028, 31571001, 31400842, 61572258, 81671785), and by the Project Sponsored by the Scientific Research Foundation for the Returned Overseas Chinese Scholars, State Education Ministry, by the Qing Lan Project and the '333' project (No. BRA2015288).